\pdfoutput=1
\documentclass[conference]{IEEEtran}
\IEEEoverridecommandlockouts
\usepackage{cite}
\usepackage{amsmath,amssymb,amsfonts}
\usepackage{algorithmic}
\usepackage{graphicx}
\usepackage{textcomp}
\usepackage{xcolor}
\def\BibTeX{{\rm B\kern-.05em{\sc i\kern-.025em b}\kern-.08em
    T\kern-.1667em\lower.7ex\hbox{E}\kern-.125emX}}
\begin{document}
\bstctlcite{BSTcontrol}

\title{Cell Instance Segmentation via\\ Multi-Task Image-to-Image Schr\"odinger Bridge\\
}

\author{
    \IEEEauthorblockN{Hayato Inoue, Shota Harada, Shumpei Takezaki and Ryoma Bise}
    \IEEEauthorblockA{
        \textit{Dept. of Information Science and Technology} \\
        \textit{Kyushu University},
        Fukuoka, Japan \\
        \{hayato.inoue@human., shumpei.takezaki@human., harada@, bise@\}ait.kyushu-u.ac.jp
    }
}
\maketitle

\begin{abstract}
Existing cell instance segmentation pipelines typically combine deterministic predictions with post-processing, which imposes limited explicit constraints on the global structure of instance masks.
In this work, we propose a multi-task image-to-image Schr\"odinger Bridge framework that formulates instance segmentation as a distribution-based image-to-image generation problem.
Boundary-aware supervision is integrated through a reverse distance map, and deterministic inference is employed to produce stable predictions.
Experimental results on the PanNuke dataset demonstrate that the proposed method achieves competitive or superior performance without relying on SAM pre-training or additional post-processing.
Additional results on the MoNuSeg dataset show robustness under limited training data.
These findings indicate that Schr\"odinger Bridge-based image-to-image generation provides an effective framework for cell instance segmentation.
\end{abstract}


\section{Introduction}

Cell instance segmentation is a fundamental task in biomedical image analysis and plays a critical role in applications such as quantitative pathology, drug discovery, and cell-level phenotyping.
Beyond foreground--background separation, this task requires accurate delineation of individual cell instances in densely packed microscopic images, where complex morphology and background artifacts often make instance separation challenging.

Most existing cell instance segmentation frameworks~\cite{cellvit, hover-net, stardist, cellpose, mediar} rely on post-processing to separate adjacent cells.
Typically, a CNN predicts intermediate dense representations (e.g., cell probability maps, distance maps, or vector fields), and individual instances are obtained through algorithmic post-processing such as watershed transforms, clustering, or morphological operations.
While effective in many settings, these post-processing pipelines often depend on dataset characteristics and hyperparameter choices, which may limit robustness and generalizability across datasets.

\begin{figure}[t]
    \centering
    \includegraphics[width=\linewidth]{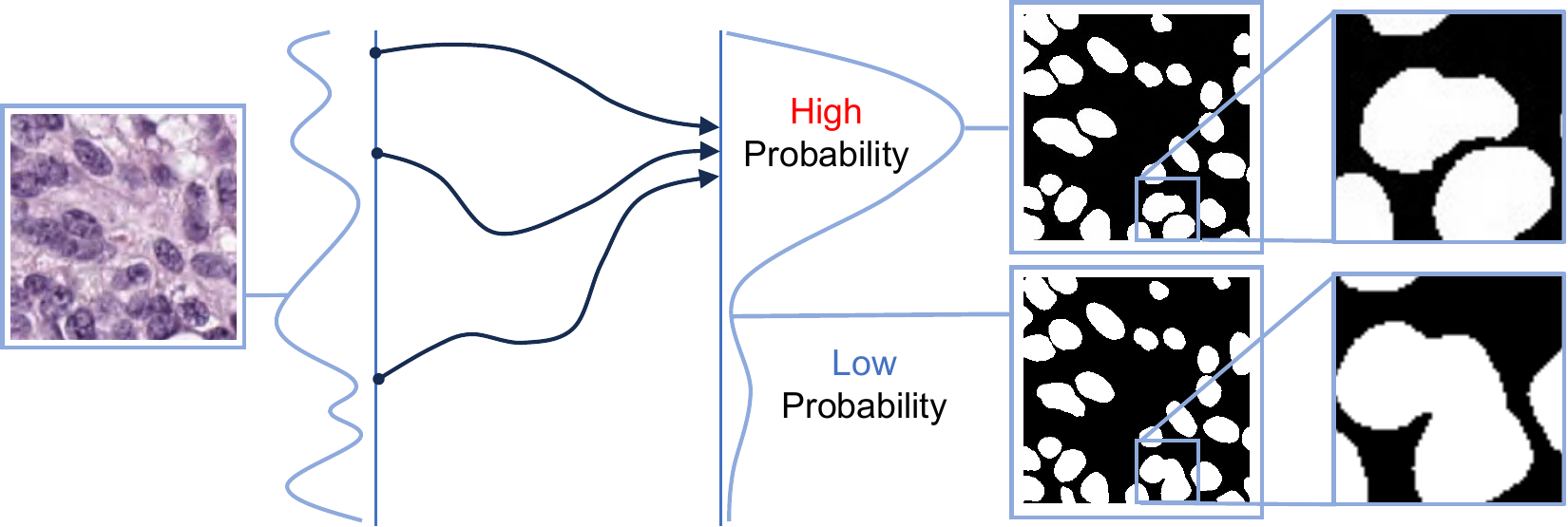}
    \caption{Conceptual illustration of distribution-based instance mask modeling.
    For a given input image, multiple candidate instance masks may exist, while only a subset corresponds to morphologically plausible cell instances.}
    \label{fig:intro_motivation}
\end{figure}

Beyond the reliance on post-processing, these pipelines are fundamentally based on deterministic predictions that impose no explicit constraints on the global structure of instance masks.
As a result, morphologically implausible outputs, such as merged cells or spurious fragments, can be produced and are commonly addressed through post-processing.
This observation motivates a shift toward \emph{distribution-based modeling}, in which valid instance masks are generated from a learned distribution that implicitly encodes morphological plausibility.
Figure~\ref{fig:intro_motivation} conceptually illustrates this motivation, highlighting that instance segmentation can be viewed as selecting morphologically plausible masks from a distribution rather than producing a single deterministic prediction.

Recent advances in generative modeling provide promising tools for learning complex data distributions.
For example, diffusion models~\cite{Rombach_2022_CVPR} can generate realistic images by iteratively denoising random noise conditioned on high-level prompts such as text descriptions or bounding boxes, effectively capturing the distribution of training data.

In contrast, cell instance segmentation requires modeling fine-grained image characteristics, including subtle cell boundaries and detailed morphological variations.
Such tasks require formulations that preserve dense spatial correspondence between the input image and the output mask, rather than relying on coarse or abstract prompts.

From this perspective, we propose a multi-task image-to-image Schr\"odinger Bridge (SB) framework for cell instance segmentation.
As a generative model, the SB~\cite{i2sb} learns a probabilistic distribution of valid outputs, which discourages the generation of morphologically implausible cell shapes.
Moreover, by formulating segmentation as an image-to-image transformation, the SB framework preserves fine-grained spatial correspondence between the input image and the output mask.
Building on this formulation, we further introduce a boundary-aware multi-task design that jointly predicts cell instance masks and a reverse distance map.

This auxiliary supervision encourages the model to capture boundary information during training, which is reflected in experimental results demonstrating consistently improved instance separation.

The main contributions of this work are summarized as follows:
\begin{itemize}
    \item We formulate cell instance segmentation as a \emph{distribution-based image-to-image generation problem}, reducing reliance on heuristic post-processing.
    \item We introduce an \emph{image-to-image SB framework} that preserves spatial correspondence while generating morphologically plausible instance masks.
    \item We propose a \emph{multi-task reverse distance map learning strategy} within the SB framework to improve instance separation.
    \item We demonstrate through experiments on public benchmarks that the proposed method achieves competitive performance across both large-scale and limited-data settings.
\end{itemize}

\section{Related Work}
\subsection{Cell Instance Segmentation}

Deep learning-based cell instance segmentation methods are commonly categorized into top-down and bottom-up approaches.
Top-down methods perform instance segmentation after object detection, typically by predicting masks within bounding boxes.
Representative examples are based on Mask R-CNN~\cite{maskrcnn}, with adaptations such as Cyto R-CNN~\cite{CytoR-CNN} for microscopic images.
While effective for isolating individual objects, these approaches are constrained by predefined bounding boxes, which can limit their performance in densely packed regions.

Bottom-up approaches instead predict pixel-wise intermediate representations and obtain instance masks through algorithmic post-processing.
For example, StarDist~\cite{stardist} predicts radial distance maps to model star-convex cell shapes, while Cellpose~\cite{cellpose} learns vector fields pointing toward cell centers and applies flow-based clustering.
More recently, CellViT~\cite{cellvit} incorporates the Segment Anything Model (SAM) as a pre-trained backbone to produce such intermediate representations, which are combined with distance-map-based post-processing to achieve strong performance.
These methods refine intermediate representations using distance maps or vector fields to improve instance separation.

A common characteristic of both top-down and bottom-up paradigms is their reliance on post-processing to convert intermediate predictions into final instance masks.
Such pipelines typically involve operations such as watershed transforms, morphological processing, and thresholding, and often require dataset-specific hyperparameter tuning, which can affect robustness and generalizability.

\subsection{Generative Models for Image-to-Image Transformation}

Generative models have been explored for modeling complex output distributions in various image analysis tasks.
In medical image analysis, diffusion-based approaches have been applied to restoration and segmentation by generating outputs from noise conditioned on input images, with a primary focus on organ- or region-level structures in modalities such as X-ray, CT, and MRI~\cite{Aimon_2023_CVPR,Huo_Ouyang_Ourselin_Sparks_2025,QiuRiy_Accurate_MICCAI2025}.

Cell instance segmentation differs from these settings in that it requires fine-grained spatial correspondence between the input image and the output mask, as well as reliable separation of densely packed instances.
While generative formulations have shown effectiveness at coarser scales, their application to such fine-grained segmentation problems remains relatively limited.

The SB provides an image-to-image generative formulation that directly relates input and output distributions under matched spatial dimensions.
Recent work, such as Image-to-Image Schr\"odinger Bridge (I\textsuperscript{2}SB)~\cite{i2sb}, has demonstrated its effectiveness for structure-preserving image-to-image generation.
In this work, we adopt this formulation and extend it with a boundary-aware multi-task design to address the requirements of cell instance segmentation.

\section{Multi-Task Schr\"odinger Bridge for Cell Instance Segmentation}

\begin{figure}[t]
    \centering
    \includegraphics[width=80mm]{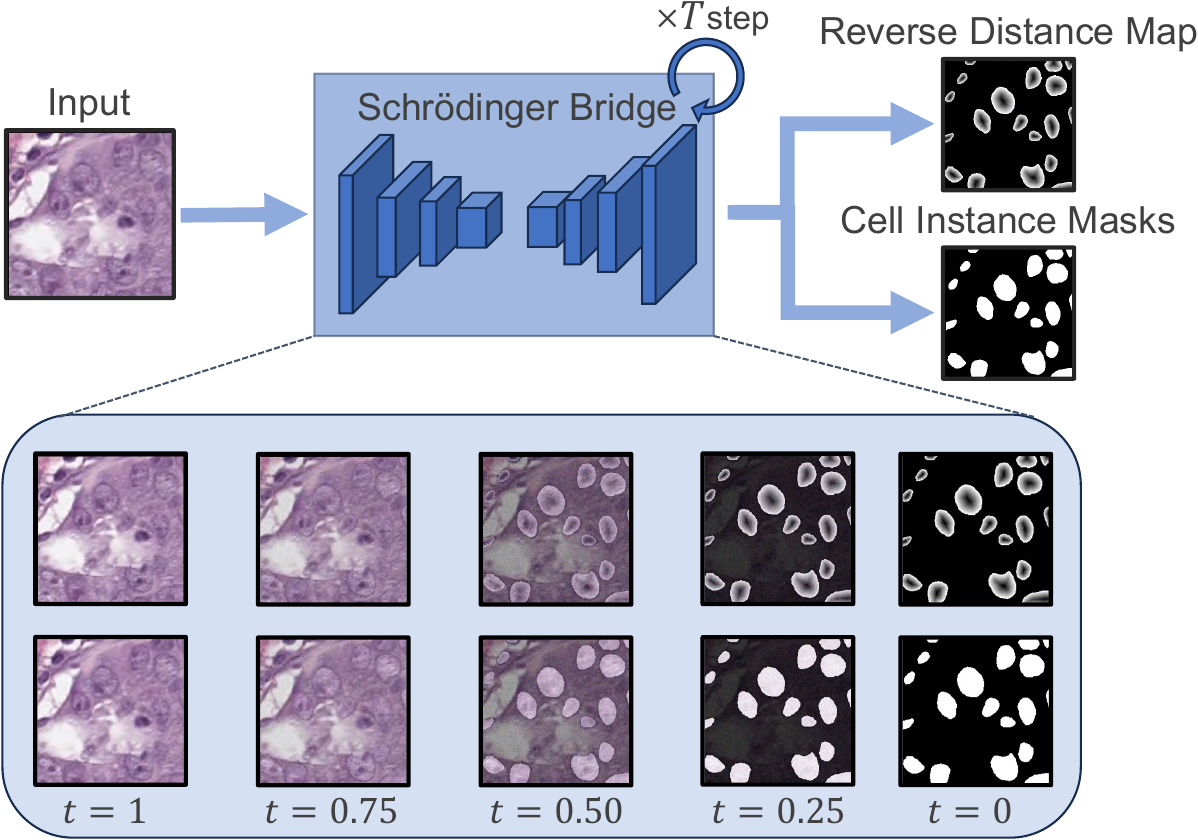}
    \caption{Overview of the proposed Multi-Task Image-to-Image SB framework.}
    \label{fig:overview}
\end{figure}

\subsection{Overview}

We consider a supervised cell instance segmentation setting with paired training data $\{(X^i, M^i)\}_{i=1}^N$, where $X^i$ is a microscopic image and $M^i$ is its ground-truth instance mask.
Given an input image $X^i$, the goal is to predict an instance mask $\hat{M}^i$ in which each connected component corresponds to an individual cell.
In ground-truth masks, cell boundaries are labeled as background, and densely packed cells are often represented as contiguous regions, making instance separation challenging.

To address this challenge, we formulate cell instance segmentation as a distribution-based image-to-image generation task using a SB.
As illustrated in Fig.~\ref{fig:overview}, the proposed framework directly models the conditional distribution of valid instance masks given an input image, eliminating the need for heuristic post-processing.
To further enhance boundary awareness, the model is trained in a multi-task manner to jointly predict the instance mask and an auxiliary reverse distance map, while only the instance mask is used at inference time.

\subsection{Reverse Distance Map for Boundary-aware Learning}

Accurate cell instance segmentation requires reliable separation of adjacent cells, particularly in densely packed regions.
In ground-truth instance masks, boundaries between neighboring cells are labeled as background, providing only weak supervision for instance separation when training with binary masks alone.

To compensate for this limitation, we introduce a \emph{reverse distance map} as an auxiliary learning target that explicitly emphasizes cell boundaries.
Unlike conventional distance maps that are often used as intermediate outputs for post-processing, the reverse distance map in our framework is used exclusively during training to provide boundary-aware supervision.

Given a ground-truth instance mask $M^i$, we apply the Euclidean Distance Transform (EDT) to each cell instance to compute the distance from each pixel to the nearest instance boundary.
The distances are normalized and inverted so that pixels closer to boundaries have higher values.
The reverse distance maps of all instances are aggregated to obtain the final map $R^i$.

By jointly predicting the instance mask and the reverse distance map, the proposed method learns boundary-aware representations that facilitate instance separation.
This auxiliary supervision is seamlessly integrated into the image-to-image SB framework while preserving a post-processing-free inference pipeline.

\subsection{Multi-Task Image-to-Image Schr\"odinger Bridge}

To generate cell instance masks in a post-processing-free manner, we adopt an I\textsuperscript{2}SB formulation that models a stochastic transport process between the input image domain and the output label domain.
This formulation directly connects the two endpoint distributions, enabling structure-preserving image-to-image generation suitable for dense instance segmentation.

The SB framework requires the input and output domains to have the same dimensionality.
In our setting, the input image $X^i \in \mathbb{R}^{H \times W \times 3}$ and the output modalities—the instance mask $M^i$ and the reverse distance map $R^i$—are aligned using a simple channel-duplication strategy.
Specifically, the input state is defined as
$\mathbf{X}_1 = (X^i, X^i) \in \mathbb{R}^{H \times W \times 6}$,
and the target state is defined as
$\mathbf{X}_0 = (M^i, M^i, M^i, R^i, R^i, R^i) \in \mathbb{R}^{H \times W \times 6}$.
Under this construction, $\mathbf{X}_1$ represents the observed image state, while $\mathbf{X}_0$ represents the target label state to be generated.

Following the I\textsuperscript{2}SB formulation~\cite{i2sb}, we define a diffusion process that directly connects $\mathbf{X}_0$ and $\mathbf{X}_1$.
Given paired samples $(\mathbf{X}_0, \mathbf{X}_1)$ during training, the posterior distribution of the intermediate state $\mathbf{X}_t$ can be derived analytically as
\begin{equation}
q(\mathbf{X}_t \mid \mathbf{X}_0, \mathbf{X}_1) = \mathcal{N}(\mathbf{X}_t; \boldsymbol{\mu}_t, \boldsymbol{\Sigma}_t),
\end{equation}
with
\begin{align}
\boldsymbol{\mu}_t =
\frac{\bar{\sigma}_t^2}{\bar{\sigma}_t^2 + \sigma_t^2}\mathbf{X}_0
+ \frac{\sigma_t^2}{\bar{\sigma}_t^2 + \sigma_t^2}\mathbf{X}_1,~
\boldsymbol{\Sigma}_t =
\frac{\sigma_t^2 \bar{\sigma}_t^2}{\bar{\sigma}_t^2 + \sigma_t^2}\mathbf{I}.
\end{align}

During training, a time $t$ is uniformly sampled from $[0,1]$, and $\mathbf{X}_t$ is sampled from the posterior distribution.
A neural network $\epsilon_\theta(\mathbf{X}_t, t)$ is trained to predict the posterior score (or equivalently, the noise term) that drives $\mathbf{X}_t$ toward $\mathbf{X}_0$ using
\begin{equation}
\mathcal{L} =
\left\|
\epsilon_\theta(\mathbf{X}_t, t) -
\frac{\mathbf{X}_t - \mathbf{X}_0}{\sigma_t}
\right\|^2
\end{equation}

This training procedure enables structure-preserving transport from the input image to the joint output of instance masks and reverse distance maps, facilitating accurate instance separation without heuristic post-processing.

\subsection{Deterministic Inference}

Although the proposed method learns a stochastic transport process between distributions during training, inference in cell instance segmentation requires a single, spatially precise prediction.
Therefore, we adopt a deterministic inference strategy that follows the most probable trajectory of the learned SB.

Specifically, during inference, random noise injection is disabled and the reverse diffusion process deterministically traces the mean of the posterior distribution from the input state to the target state.
This design avoids stochastic fluctuations that may blur boundaries or cause spatial misalignment, yielding stable and reproducible instance masks.
The auxiliary reverse distance map is generated as part of the process but is discarded at inference time.
The final cell instance mask is obtained by averaging the three output mask channels and applying a fixed threshold of $0.5$ to the resulting map, which serves as a deterministic decision rule analogous to the argmax operation commonly used in conventional segmentation networks, rather than a heuristic post-processing step.

\section{Experiments}
\subsection{Datasets}

We evaluate the proposed method under two complementary settings: a large-scale training regime and a limited training data regime.
These experiments are designed to assess overall instance segmentation performance, robustness to data scarcity, and the role of post-processing across different datasets.

\textbf{PanNuke}~\cite{pannuke1,pannuke2} is a large-scale histopathology dataset for cell instance segmentation, consisting of 7,904 image patches of size $256 \times 256$ pixels with annotations for 189,744 cells across 19 tissue types.
This dataset represents a data-rich setting and is used to evaluate performance under large-scale training data conditions. 

\textbf{MoNuSeg}~\cite{kumar2017dataset,kumar2019multi} is a nucleus instance segmentation dataset with a substantially smaller number of annotated images.
It contains 30 training images and a predefined test set consisting of 14 images derived from H\&E-stained tissue samples acquired at $40\times$ magnification, with annotations for approximately 22,000 nuclei across seven organs.
This dataset represents a data-limited setting and is used to evaluate robustness under scarce training data.

In all experiments, cells are treated as a single class to focus on instance separation rather than cell-type classification.

\begin{table}[t]
  \centering
  \small
  \caption{Quantitative comparison on the PanNuke dataset under the large-scale training setting.}
  \label{tab:result}
  \begin{tabular}{l|cccc}
    \hline \hline
    Method & bPQ $\uparrow$ & F1 $\uparrow$ & Recall $\uparrow$ & Precision $\uparrow$ \\
    \hline
    U-Net & 0.5799 & 0.7930 & 0.7989 & 0.7874  \\
    \hline
    U-Net+R & 0.5837 & 0.7942 & 0.7972 & 0.7908 \\
    \hline
    CellViT & 0.6149 & 0.8037 & \textbf{0.8328} & 0.7984 \\
    \hline
    CellViT+proc & 0.6221 & 0.7946 & 0.7847 & 0.8357  \\
    \hline
    \textbf{Ours} & \textbf{0.6362} & \textbf{0.8106} & 0.8011 & \textbf{0.8458} \\
    \hline \hline
    
  \end{tabular}
\end{table}

\subsection{Comparative Methods}

We compare the proposed method with deterministic baselines and a state-of-the-art instance segmentation framework to clarify the effects of boundary-aware supervision, generative modeling, and post-processing.

As deterministic baselines, we employ a standard U-Net trained only with binary instance masks (U-Net) and its multi-task variant trained with both instance masks and the proposed reverse distance maps (U-Net+R).
Comparing these models isolates the effect of boundary-aware supervision, while comparing U-Net+R with the proposed method highlights the benefit of distribution-based generative modeling under the same input information.

We also compare with CellViT~\cite{cellvit}, a state-of-the-art Transformer-based framework for cell instance segmentation that incorporates the Segment Anything Model (SAM) as a pre-trained backbone.
CellViT demonstrated that leveraging SAM pre-training can substantially improve performance in cell instance segmentation.
In addition, CellViT employs distance-map prediction followed by a dedicated post-processing pipeline for instance separation.
To assess the role of post-processing, we evaluate CellViT both without post-processing (CellViT) and with its full post-processing pipeline (CellViT+proc).

\subsection{Experimental Settings}

For CellViT, we follow the training protocol reported in the original paper, including initialization with SAM-H.
Baseline U-Net models are trained using the Adam optimizer with a learning rate of $1 \times 10^{-3}$ and early stopping based on validation loss.
For U-Net+R, the loss is defined as the sum of binary cross-entropy for the instance mask and mean squared error for the reverse distance map.
Note that the reverse distance maps were pre-computed for all training samples.

The proposed method employs a U-Net backbone initialized with an unconditional ADM pretrained on ImageNet.
Training is performed for 300{,}000 iterations using Adam with an initial learning rate of $5 \times 10^{-5}$.
An exponential moving average (EMA) of model parameters is applied during inference, and the reverse diffusion process uses $T=50$ steps with a symmetric noise schedule ($\beta_{\mathrm{max}}=0.3$).

\subsection{Evaluation Metrics}

We evaluate instance segmentation performance using binary Panoptic Quality (bPQ), which jointly measures segmentation and detection accuracy.
Since all cells are treated as a single class, bPQ is computed as the product of Segmentation Quality (SQ) and Detection Quality (DQ).

We additionally report Precision, Recall, and F1-score based on centroid matching.
A predicted instance is considered a true positive if its centroid lies within a 12-pixel radius of a ground-truth centroid, corresponding approximately to the typical cell radius in the evaluated datasets.

\begin{table}[t]
  \centering
  \small
  \caption{Ablation study on the PanNuke dataset under the large-scale training setting.}
  \label{tab:Ablation_study}
  \begin{tabular}{l|cccc}
    \hline
    Method & bPQ $\uparrow$ & F1 $\uparrow$ & Recall $\uparrow$ & Precision $\uparrow$ \\
    \hline
    Single-Task Mask & 0.5044 & 0.6815 & 0.6790 & 0.7304 \\
    Single-Task RVdist & 0.5903 & 0.7756 & 0.7460 & 0.8373 \\
    \hline
    \textbf{Multi-Task} & \textbf{0.6341} & \textbf{0.8100} & \textbf{0.8017} & \textbf{0.8448} \\
    \hline
  \end{tabular}
\end{table}

\begin{figure}[t]
  \centering
  \includegraphics[width=90mm]{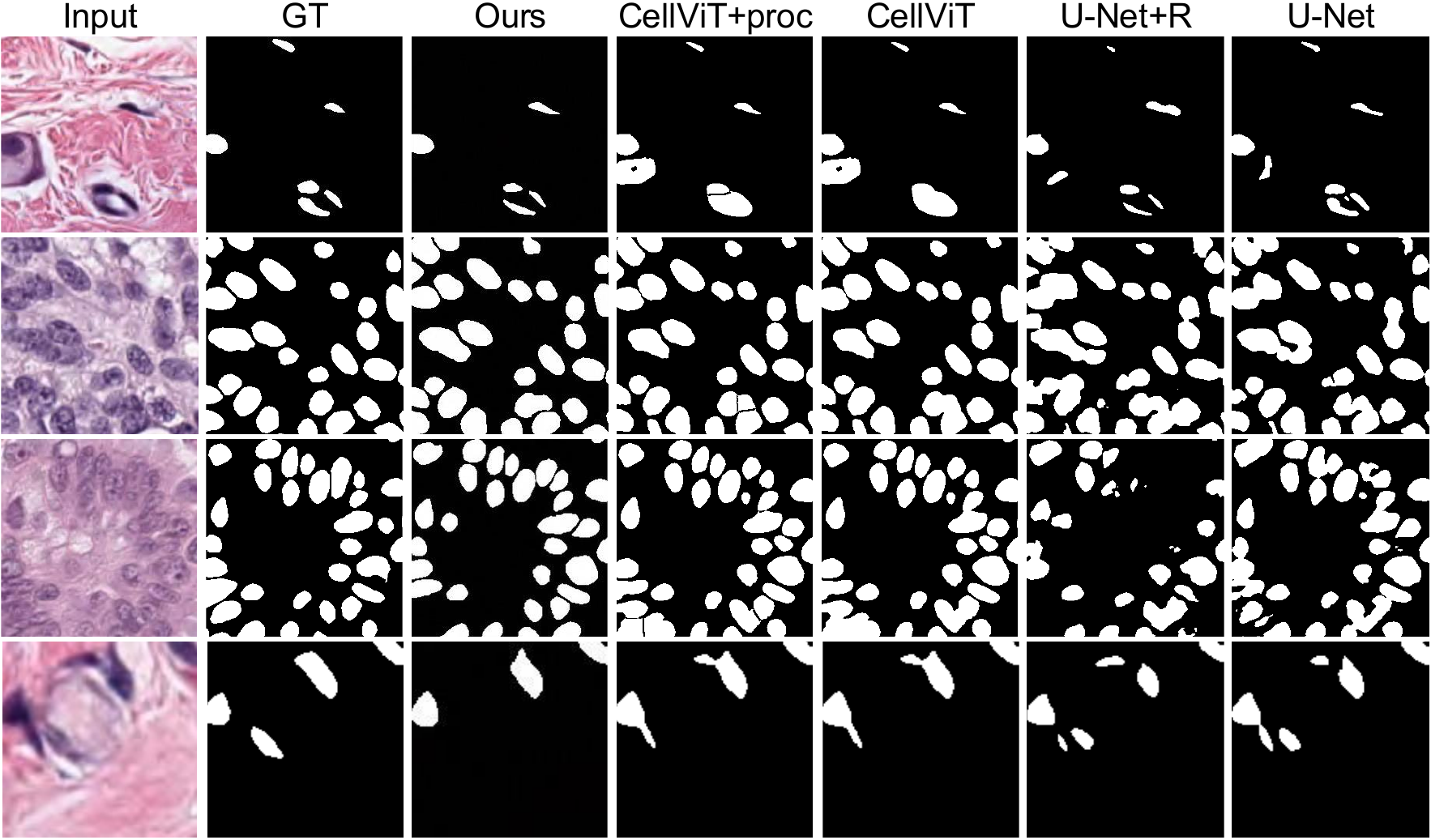}
  \caption{Qualitative cell instance segmentation results on the PanNuke dataset.}
  \label{fig:result_image}
\end{figure}

\subsection{Performance under Large-scale Training Data}

\begin{figure*}[t]
    \centering
    \includegraphics[width=1.0\textwidth]{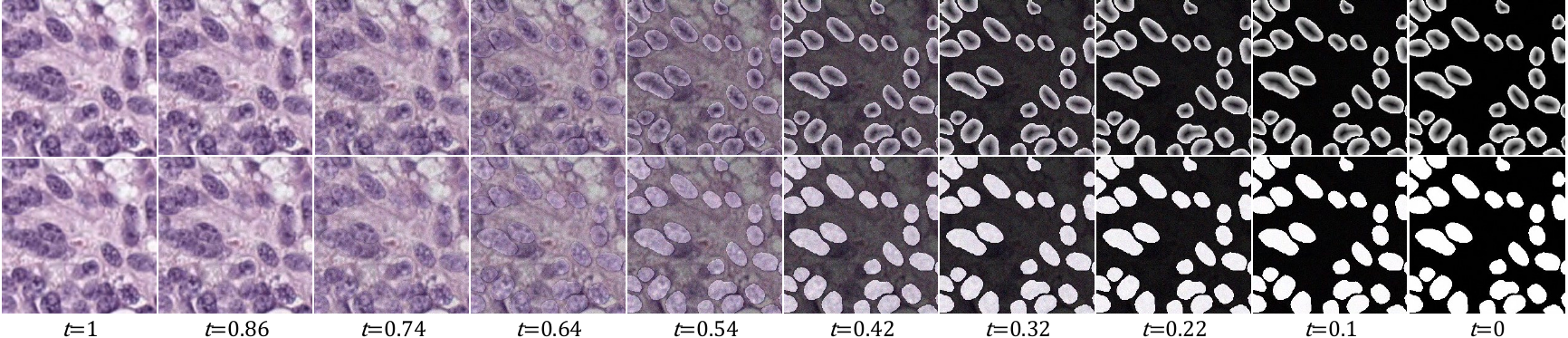}
    \caption{Transformation process at time step.}
    \label{fig:SB_process}
\end{figure*}

We first evaluated the proposed method under a large-scale training setting to assess whether the proposed post-processing-free approach can achieve competitive instance segmentation accuracy when sufficient annotated data are available.
For this purpose, we use the PanNuke dataset.
We adopt a 3-fold cross-validation protocol to obtain stable performance estimates.
The dataset is randomly split into three folds, and for each fold, one split is used for training, one for validation, and one for testing.
Model selection and early stopping are performed using the validation split, and all reported results are averaged over the three folds.

Table~\ref{tab:result} reports the quantitative results on the PanNuke dataset under the large-scale training setting.
Among the deterministic baselines, U-Net+R consistently outperforms the standard U-Net, demonstrating that incorporating reverse distance map supervision improves instance segmentation performance even without generative modeling.

Compared with U-Net-based baselines, CellViT without post-processing achieves a substantial performance gain, indicating the effectiveness of SAM pre-training for cell instance segmentation.
This result highlights the strong impact of leveraging large-scale pre-trained models in this task.
Applying post-processing further improves the performance of CellViT, confirming that heuristic post-processing can enhance instance separation accuracy in a data-rich setting.

Notably, the proposed method achieves the highest bPQ, F1-score and Precision among all compared methods, despite not using SAM pre-training or post-processing.
These results demonstrate that the proposed method can surpass or match state-of-the-art performance through a unified generative framework without relying on external pre-training or dataset-specific post-processing.

To better understand which components contribute to this performance gain, we further analyze the proposed design through an ablation study on the PanNuke dataset.
The results are summarized in Table~\ref{tab:Ablation_study}.
All ablation variants are implemented within the same SB framework to ensure a fair comparison.
We investigate three variants of the proposed framework:
(i) \textbf{Single-Task Mask}, which predicts only the instance mask using SB;
(ii) \textbf{Single-Task RVdist}, which predicts only the reverse distance map using SB, from which a cell instance mask is obtained via a simple hole-filling operation without additional post-processing; and
(iii) \textbf{Multi-Task}, which jointly predicts both outputs within the proposed multi-task SB formulation (our method).

As shown in Table~\ref{tab:Ablation_study}, single-task learning with either the instance mask or the reverse distance map alone does not achieve optimal performance.
While predicting the reverse distance map alone performs better than using the instance mask alone, the best performance is obtained when both outputs are learned jointly.
The multi-task formulation consistently improves all evaluation metrics, indicating that boundary-aware supervision plays a critical role in accurate instance separation.

These component-wise analyses help explain the superior performance of the proposed method and are further supported by qualitative results in Fig.~\ref{fig:result_image}.
Deterministic baselines and CellViT often produce merged instances or spurious detections in dense regions, even with post-processing.
In contrast, the proposed method separates adjacent cells with morphologically plausible shapes and suppresses false positives without post-processing, consistent with the quantitative results.

Measured on an NVIDIA RTX A6000 GPU, our method inference requires 3.5~s/patch ($T=50$) compared to CellViT's 0.35~s. However, this trade-off is justified by superior performance and the elimination of complex post-processing.

\subsection{Performance under Limited Training Data (MoNuSeg)}

Table~\ref{tab:monuseg_result} reports the quantitative results on the MoNuSeg dataset.
This experiment evaluates the robustness of the proposed method under a limited training data setting, where only a small number of annotated images are available.

In this data-scarce scenario, SAM pre-training provides a clear advantage, as evidenced by the substantial performance gap between CellViT and U-Net-based baselines.
This trend is more pronounced than in the PanNuke experiments, indicating that large-scale pre-training becomes increasingly beneficial when annotated data are limited.
In addition, post-processing yields a larger improvement in bPQ on MoNuSeg than on PanNuke, suggesting that the effectiveness of post-processing is highly dataset-dependent.

Under these conditions, the proposed method demonstrates strong performance without relying on post-processing.
Specifically, without post-processing, the proposed method achieves the highest F1-score and Precision among all compared methods, indicating accurate nucleus detection with reduced over-detection.
Despite the advantages of SAM pre-training and post-processing for CellViT, the proposed method attains a bPQ comparable to CellViT with post-processing, while outperforming it in terms of F1-score.

\begin{table}[tb]
  \centering
  \small
  \caption{Quantitative comparison on the MoNuSeg dataset under the limited training data setting.}
  \label{tab:monuseg_result}
  \begin{tabular}{l|cccc}
    \hline \hline
    Method & bPQ $\uparrow$ & F1 $\uparrow$ & Recall $\uparrow$ & Precision $\uparrow$ \\
    \hline
    U-Net & 0.5066 & 0.8051 & 0.8426 & 0.7707 \\
    \hline
    U-Net+R & 0.5335 & 0.8098 & 0.8541 & 0.7698 \\
    \hline
    CellViT & 0.5842 & 0.8468 & \textbf{0.8891} & 0.8142 \\
    \hline
    CellViT+proc & \textbf{0.6066} & 0.8444 & 0.8506 & 0.8449 \\
    \hline
    \textbf{Ours} & 0.6017 & \textbf{0.8563} & 0.8689 & \textbf{0.8493} \\
    \hline \hline
  \end{tabular}
\end{table}

These results indicate that, even under limited training data, the proposed post-processing-free approach can achieve competitive instance segmentation performance.
Notably, despite being based on a generative diffusion framework, the proposed method does not critically rely on large-scale annotated datasets and maintains stable performance across different data regimes.

\begin{figure}[t]
    \centering
    \includegraphics[width=90mm]{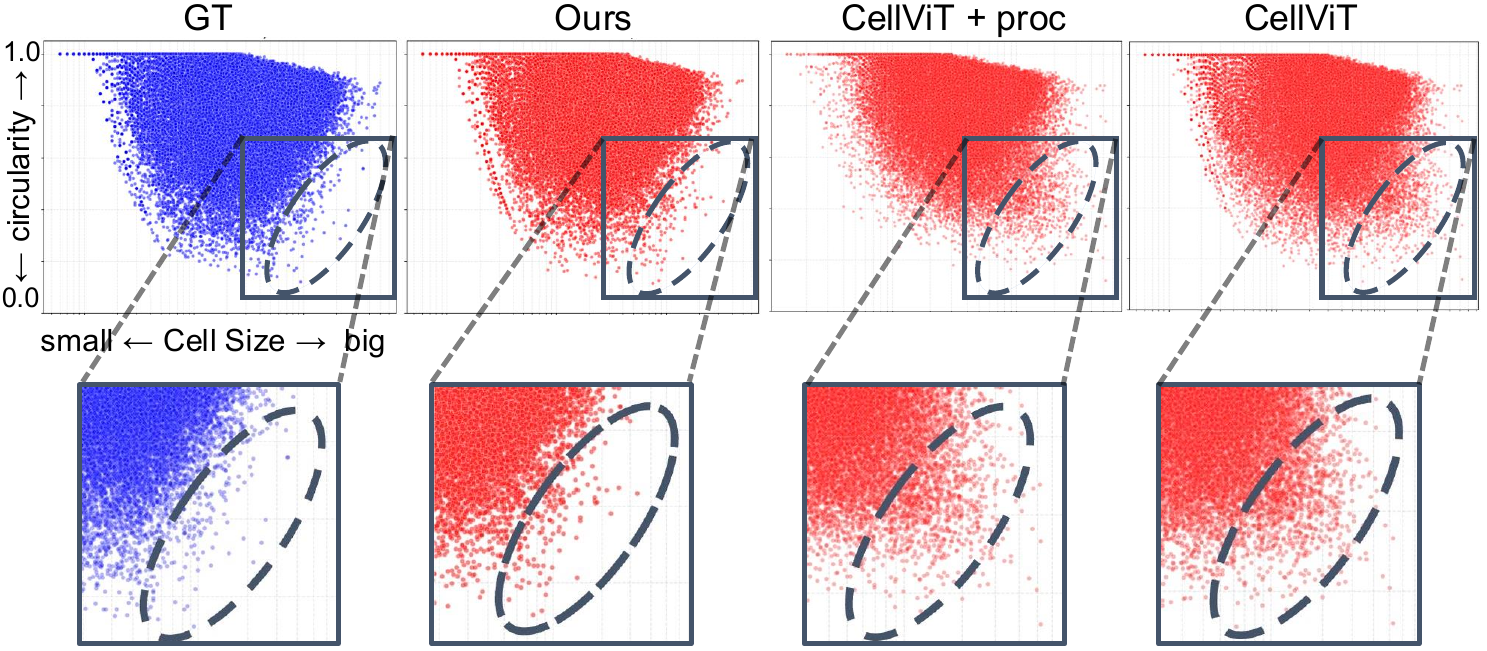}
    \caption{Joint distributions of cell size and circularity for the ground truth (GT), the proposed method, and CellViT variants on the PanNuke dataset. 
    The proposed method produces a distribution closer to the GT, suppressing morphologically atypical cell instances.}
    \label{fig:shape_analysis}
\end{figure}

\subsection{Analysis of Instance Shape Characteristics}

We analyze both the generation process and the morphological characteristics of predicted instance masks on the PanNuke dataset.
Figure~\ref{fig:SB_process} illustrates the reverse diffusion process, where cell instances gradually emerge and stabilize, indicating structure-preserving image-to-image generation.
This behavior suggests that the proposed framework progressively refines instance boundaries while preserving spatial correspondence throughout the generation process.

To examine shape characteristics, we analyze the joint distribution of cell size and circularity for the ground truth (GT), the proposed method, and CellViT variants (Fig.~\ref{fig:shape_analysis}).
The GT distribution is concentrated in a region corresponding to morphologically plausible cell shapes, while only a few samples appear in low-density regions representing atypical shapes.
In contrast, CellViT produces a noticeable number of instances in these low-density regions, indicating uncommon combinations of size and circularity, which are only partially reduced by post-processing.
The proposed method substantially suppresses such instances, resulting in a distribution that more closely matches the GT.
These results indicate that the proposed method preserves the morphological characteristics of cell instances while achieving accurate segmentation without relying on post-processing.

\section{Conclusion}

We proposed a cell instance segmentation framework based on a multi-task image-to-image SB that reduces reliance on heuristic post-processing.
By formulating instance segmentation as a distribution-based image-to-image generation problem, the proposed method generates morphologically plausible instance masks within a unified generative process.
Boundary-aware supervision is incorporated through a reverse distance map, enabling accurate instance separation with stable and spatially consistent predictions.

Experiments on the PanNuke dataset demonstrate that the proposed method achieves competitive or superior performance compared with state-of-the-art approaches, despite not relying on SAM pre-training or post-processing.
Additional results on the MoNuSeg dataset show robustness under limited training data, even when large-scale pre-training and post-processing benefit existing methods.
Furthermore, shape analysis confirms that the proposed method produces instance distributions closer to the ground truth, highlighting the effectiveness of SB-based image-to-image generation for cell instance segmentation.

\subsection*{Acknowledgment}
This work was supported by JSPS KAKENHI Grant Number JP25K22846, ASPIRE Grant Number JPMJAP2403, and SIP Grant Number JPJ012425. We used ABCI 3.0 provided by AIST and AIST Solutions with support from ``ABCI 3.0 Development Acceleration Use.''
\bibliographystyle{IEEEtran}
\bibliography{refs_short}
\end{document}